\pdfoutput=1

\documentclass[10pt,twocolumn,letterpaper]{article}

\usepackage[pagenumbers]{cvpr} 

\usepackage[dvipsnames]{xcolor}

\usepackage[accsupp]{axessibility}
\usepackage{graphicx}
\usepackage{amsmath}
\usepackage{amssymb}
\usepackage{booktabs}
\usepackage{xcolor}
\usepackage{colortbl}
\usepackage{bbding}
\usepackage{wrapfig}
\usepackage{setspace}

\usepackage{graphicx}       
\usepackage{multirow}
\usepackage{color}
\usepackage{kotex}
\usepackage{kotex-logo}
\usepackage{algorithm}
\usepackage{algpseudocode}
\usepackage{arydshln}
\usepackage{cite}

\usepackage{caption}
\usepackage{adjustbox}
\usepackage{wrapfig}
\usepackage{lipsum}
\usepackage{array}
\usepackage{makecell}
\usepackage{booktabs}

\definecolor{cvprblue}{rgb}{0.21,0.49,0.74}
\usepackage[pagebackref,breaklinks,colorlinks,citecolor=cvprblue]{hyperref}

\usepackage[capitalize]{cleveref}
\crefname{section}{Sec.}{Secs.}
\Crefname{section}{Section}{Sections}
\Crefname{table}{Table}{Tables}
\crefname{table}{Tab.}{Tabs.}

\begin{document}

 \author{Dongheon Lee\qquad Seokju Yun\qquad Youngmin Ro\thanks{Corresponding author (E-mail: \tt youngmin.ro@uos.ac.kr)}\\Machine Intelligence Laboratory, University of Seoul, Korea \\ 
\tt \small Code: \href{https://github.com/dslisleedh/PLKSR}{https://github.com/dslisleedh/PLKSR}}

\title{Partial Large Kernel CNNs for Efficient Super-Resolution}




\maketitle

\begin{abstract} 
Recently, in the super-resolution~(SR) domain, transformers have outperformed CNNs with fewer FLOPs and fewer parameters since they can deal with long-range dependency and adaptively adjust weights based on instance.
In this paper, we demonstrate that CNNs, although less focused on in the current SR domain, surpass Transformers in direct efficiency measures. 
By incorporating the advantages of Transformers into CNNs, we aim to achieve both computational efficiency and enhanced performance.
However, using a large kernel in the SR domain, which mainly processes large images, incurs a large computational overhead.
To overcome this, we propose novel approaches to employing the large kernel, which can reduce latency by 86\% compared to the naive large kernel, and leverage an Element-wise Attention module to imitate instance-dependent weights.
As a result, we introduce Partial Large Kernel CNNs for Efficient Super-Resolution (PLKSR), which achieves state-of-the-art performance on four datasets at a scale of $\times$4, with reductions of 68.1\% in latency and 80.2\% in maximum GPU memory occupancy compared to SRFormer-light.
\end{abstract}
\section{Introduction}\label{intro}
\begin{figure}[]
  \centering
  \includegraphics[width=\columnwidth]{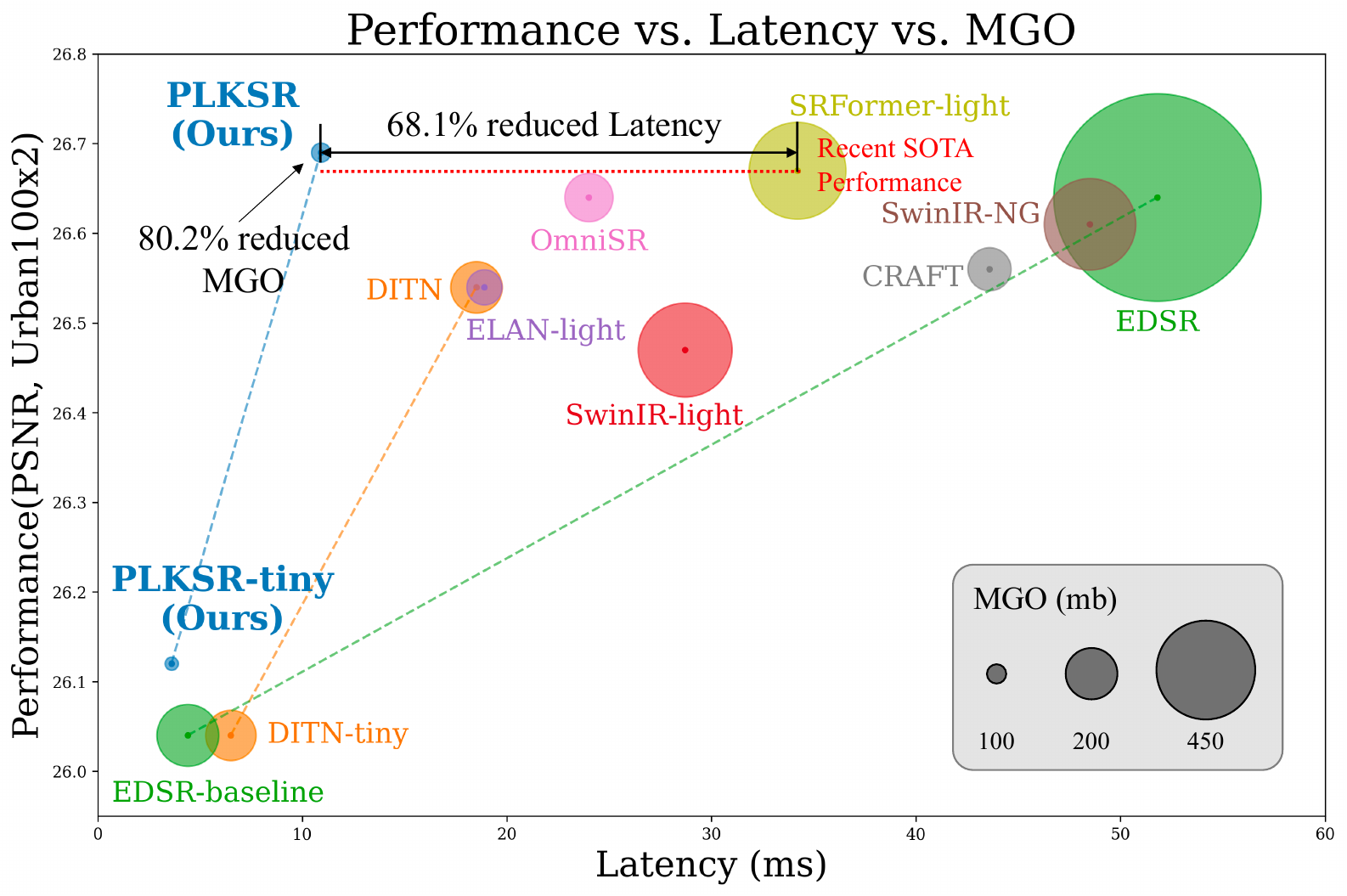}
  \vspace{-0.7cm}
  \caption{Comparison on performance, latency, and Maximum GPU memory Occupancy~(MGO) at BSD100$\times$2 dataset. Our PLKSR performs the best when compared to the SOTA SR methods, with 68.1 and 80.2 less latency and MGO compared to SRFormer-light, respectively. All metrics are measured by restoring an HD (1280×720) image using RTX4090 GPU at FP16 precision.}
  \label{fig:mainfig}
\end{figure}
Image Super-Resolution~(SR) aims to reconstruct High-Resolution~(HR) images from Low-Resolution~(LR) observations.
A fundamental challenge in SR is the ill-posed nature of the task, where a single LR image may correspond to multiple valid HR reconstructions.
The advent of Convolutional Neural Networks~(CNNs)~\cite{SRCNN, EDSR, RCAN, ABPN, HPUN, ShuffleMixer, SAFMN, PCEVA} has significantly advanced this task, thanks to their ability to process local features efficiently.
The recent rise in the use of streaming media has increased the demands to restore or enhance high-resolution images on resource-constrained high-definition devices such as smartphones, driving the need for lightweight SR models. 
Due to these demands,
Transformers~\cite{ViT, SwinT, Restormer, SwinIR, ELAN, Restormer, HAT, OmniSR, SRFormer, DAT, CRAFT, DLGSANet, DITN}, which outperform CNNs with relatively fewer FLoating point OPerations~(FLOPs) and parameter counts using Multi-Head Self-Attention~(MHSA), are emerging as a promising alternative.

\textbf{However, we observe that FLOPs and parameter counts do not consistently align with more direct efficiency measures, such as latency or Maximum GPU memory Occupancy~(MGO).}
In Table~\ref{tab:Discrepancy}, we compare the FLOPs, parameter counts, latency, and MGO required by representative CNNs~\cite{EDSR, RCAN} and Transformers~\cite{SwinIR, SRFormer} to process a single image.
Surprisingly, despite having on average 8.1$\times$ and 9.5$\times$ as many FLOPs and parameter counts, respectively, CNNs actually outperform Transformers in efficiency, exhibiting on average 2.4$\times$ lower latency and 4.9$\times$ lower maximum GPU memory occupancy.
Therefore, this study aims to enhance a CNN-based model by incorporating the factors that contribute to Transformers' superior performance while maintaining computational efficiency.

Transformers are known to outperform CNNs because they can handle long-range dependencies and generate instance-dependent weights\cite{SwinIR}.
Through rigorous exploration and experiments, we develop an efficient CNN-based model that successfully integrates the Transformer's capabilities for handling long-range dependencies and providing instance-dependent weighting.
Firstly, to improve efficiency, our approach focuses on processing only a specified chunk of input features. While previous methods~\cite{FasterNet, SHViT} lacked clear criteria for dividing the chunk, our rigorous experiments suggest optimal criteria for this division in terms of latency. 
Although it is common to enlarge the receptive field of CNNs by stacking multiple small convolutional kernels~\cite{VGGNet}, our investigation demonstrates that using a single large kernel offers faster latency with similar MGO levels compared to stacking small kernels multiple times.
Furthermore, our visual analysis (see in Figure~\ref{fig:frequency} and Figure~\ref{fig:featuremap_vis}) indicates that employing a single large kernel aligns more closely with the Transformer's ability to capture low-frequency features, such as edges and structures, compared to the use of multiple small kernels.
Finally, while dynamic convolution offers a straightforward way of generating instance-dependent weights, our experimental findings indicate that it adversely affects latency. Thus, we leverage Element-wise Attention (EA) into our model, which assigns individual attention weights to each element within the input feature tensor, optimizing performance.

Drawing on these insights, we introduce Partial Large Kernel CNNs for Efficient Super-Resolution~(PLKSR).
Our PLKSR utilizes advantages from both CNNs and Transformers, achieving state-of-the-art performance on four datasets at scale $\times$4 while achieving 42.3\% lower latency and 45.6\% lower MGO than ELAN-light~\cite{ELAN}.
The tiny variants of PLKSR maintain high performance even when scaled down, demonstrating superior efficiency compared to other approaches that utilize large receptive fields.
The ablation study confirms that each proposal in our model significantly contributes to performance improvements.
Compared to tiny variants of PLKSR (PLKSR-tiny) and other approaches with large receptive fields on an edge device (iPhone 12), Our PLKSR-tiny demonstrates the lowest latency, highlighting its superior efficiency.
Additionally, by visualizing the pixels used by the model to reconstruct images, we demonstrate that compared to other methods, PLKSR effectively utilizes the long-range dependencies captured by its large kernels.

Our contributions can be summarized as follows:
\begin{itemize}
    \item We develop a CNN-based model incorporating the Transformer's capabilities for handling long-range dependencies and instance-dependent weighting, optimizing processing and division of input features for enhanced efficiency.
    \item Our proposed model achieves state-of-the-art performance on four datasets with significant reductions in latency and MGO.
    \item We demonstrate through visual analysis and empirical testing that PLKSR effectively captures essential low-frequency features for super-resolution, similar to transformers.
\end{itemize}

\section{Related Work}
\subsection{Super-Resolution Models}
Since SRCNN~\cite{SRCNN} has shown better performance than existing super-resolution methods, various researches~\cite{EDSR, RCAN, PAN, ABPN, PCEVA, HPUN} have proposed super-resolution models using CNNs.
Recently, Transformers~\cite{SwinIR, Restormer} outperformed CNNs with lower FLOPs and parameter counts due to their ability to handle long-range dependencies.
Many researchers have enhanced Transformer by widening the receptive field ~\cite{SRFormer, HAT}, extracting various features ~\cite{OmniSR, CRAFT, DAT}, or speeding up inference speed by removing layer normalization and proposing simplified MHSA~\cite{ELAN, DITN}.
\subsection{Large kernel CNNs}
Since VGGNet~\cite{VGGNet}, most CNNs have used 3$\times$3 kernel convolution, but recent studies reported comparable performance to Transformer by using large kernel convolution in CNNs.
For example, ConvNeXt~\cite{ConvNeXt} performed comparably to Swin Transformer~\cite{SwinT} by leveraging a 7$\times$7 depth-wise convolution~(DWC), and DWNet~\cite{DWNet} introduced dynamic convolution focusing on similarity of attention and DWC.
RepLKNet~\cite{RepLKNet} and SLaK~\cite{SLaK} employed inverse implicit GEneralized Matrix Multiplication~(iGEMM) DWC for their efficiency and scaled up the kernel size up to 51$\times$51.
Also, in the Super-Resolution task, there have been efforts to mimic MHSA operations with DWC~\cite{ShuffleMixer}, max pooling~\cite{SAFMN}, and dynamic convolution~\cite{DLGSANet}.
\subsection{Partial Channel Design}
FasterNet~\cite{FasterNet} proposed Partial Convolution~ (PConv), which performs convolution operations on a subset of channels.
SHViT~\cite{SHViT} reduced multi-head redundancy by computing single-head self-attention on the subset of channels.
PCEVA~\cite{PCEVA} extracted multi-scale features using PConvs sequentially with decreasing partial convolution channels.

\section{Analysis}
\subsection{Discripency Btw. Direct and Indirect Metrics}
\begin{table}[]
\caption{
    Discrepancies between FLOPs/parameter counts and latency/MGO. All metrics are measured by restoring an HD~(1280$\times$720) image at scale $\times$2 using RTX4090 GPU at FP16 precision.
}
\label{tab:Discrepancy}
\resizebox{\columnwidth}{!}{%
\begin{tabular}{@{}l|l|cc|cc@{}}
\toprule
Arch. & Methods & \begin{tabular}[c]{@{}c@{}}\#FLOPs\\ (G)\end{tabular} & \begin{tabular}[c]{@{}c@{}}\#Params\\ (K)\end{tabular} & \begin{tabular}[c]{@{}c@{}}Latency\\ (ms)\end{tabular} & \begin{tabular}[c]{@{}c@{}}MGO\\ (mb)\end{tabular} \\ \midrule
\multirow{2}{*}{CNNs} & EDSR-baseline~\cite{EDSR} & 316 & 1370 & \textbf{9.9} & \textbf{320.2} \\
 & RCAN~\cite{RCAN} & 3530 & 15444 & 101.8 & 401.2 \\ \midrule
\multirow{2}{*}{Transformers} & SwinIR-light~\cite{SwinIR} & 243.7 & 910 & 124.9 & 1764.3 \\
 & SRFormer-light~\cite{SRFormer} & \textbf{229.4} & \textbf{853} & 152.7 & 1744.4 \\ \bottomrule
\end{tabular}
}
\end{table}

Previous research has focused on Transformers for SR tasks, considering them more efficient than CNNs because they require fewer parameters and FLOPs.
However, recent studies~\cite{ShuffleNetV2, RepVGG, FasterNet} have argued that parameter counts and FLOPs may not accurately indicate a model's efficiency.
To verify this in SR tasks, we measure the parameter counts and FLOPs of representative CNN-based~\cite{EDSR, RCAN} and Transformer-based~\cite{SwinIR, SRFormer} models, along with more direct indicators such as latency and Maximum GPU memory Occupancy (MGO).
In Table~\ref{tab:Discrepancy}, EDSR-baseline has 1.4$\times$ and 1.6$\times$ more FLOPs and Parameters than SRFormer-light, which seems inefficient, but it is \textbf{15.5$\times$} faster in terms of latency and has \textbf{5.4$\times$} less MGO.
Furthermore, RCAN has 15.4$\times$ and 18.1$\times$ more FLOPs and parameter counts than SRFormer-light, respectively, but is 1.5$\times$ faster in latency and has 4$\times$ fewer on MGO.
This can be explained by the fact that the convolution module has a more efficient memory access pattern than the Multi-Head Self-Attension (MHSA) module~\cite{flashattn} and rarely uses operations that increase latency while adding little FLOPs, such as reshaping or layer normalization~\cite{ShuffleNetV2}.

\subsection{Importance of Large Kernel}
\begin{figure}[]
  \centering
  \includegraphics[width=\columnwidth]{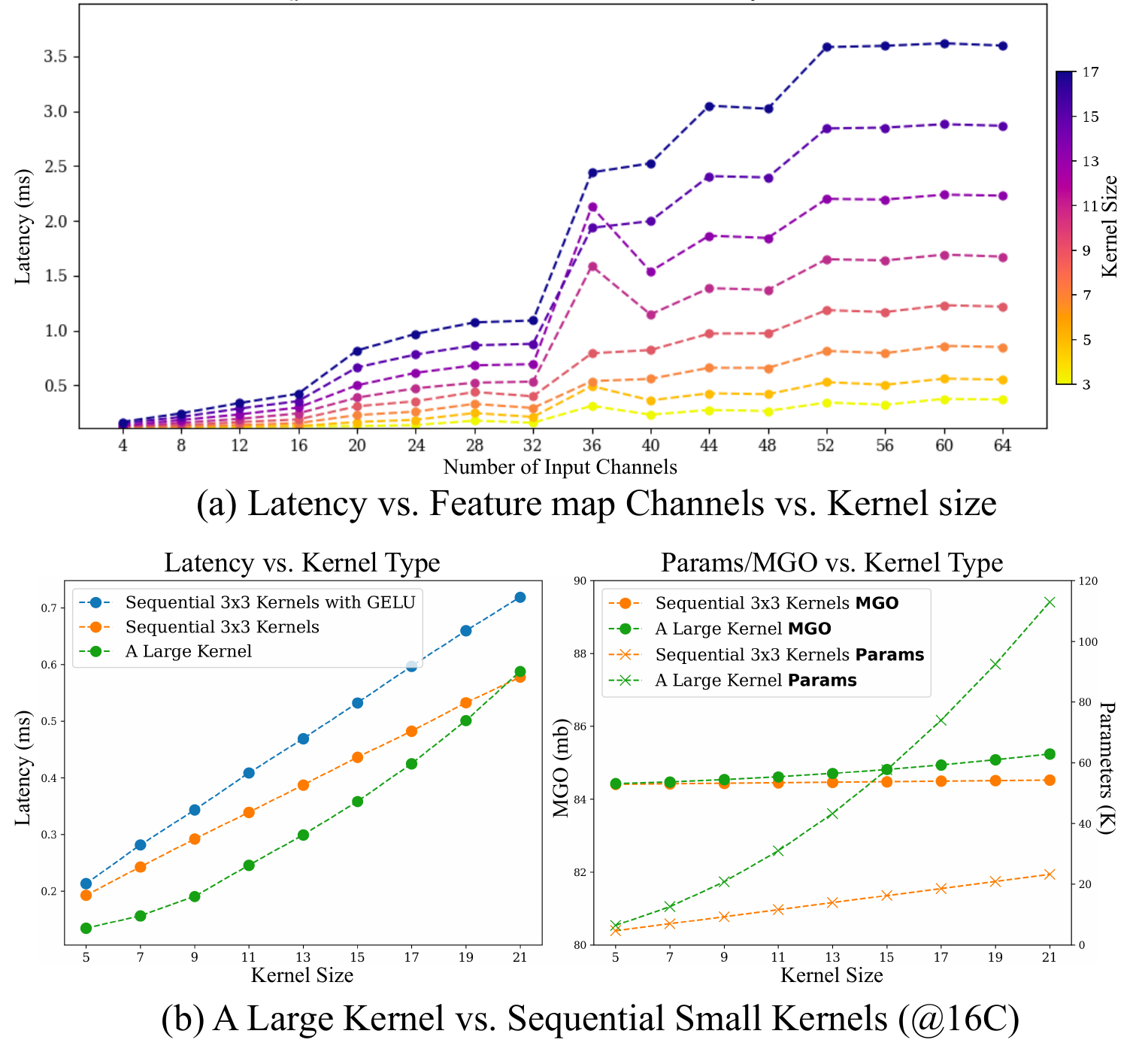}
  \vspace{-0.7cm}
  \caption{
    Analysis of convolution operation. 
    (a) demonstrates the change in latency of convolution operation as the channel and kernel size change, and (b) demonstrates the latency, MGO, and parameters of using a large kernel directly and using successive small kernels (w/ and w/o GELU) with the same receptive field as the large kernel when computing convolution on 16 channels. 
    All metrics are measured by processing a feature map with a size of 640$\times$360 using RTX4090 GPU at FP16 precision.
  }
  \label{fig:pconv}
\end{figure}

The ability to deal with long-range dependency distinguishes Transformers from traditional CNNs as a key advantage.
However, in the SR task, which primarily processes large images, simply imitating this advantage with convolution results in significant computational overhead. 
Inspired by previous studies~\cite{FasterNet, SHViT}, we address long-range dependencies in only a subset of channels.
As shown in Figure~\ref{fig:pconv} (a), we observed that latency increases are minimal up to 16 channels. 
Therefore, in our study, we handle long-range dependencies at 16 channels, and the careful selection of channels differs from previous studies.
In CNNs, implementing a large receptive field is typically achieved by utilizing either a single large kernel or stacked 3$\times$3 kernels.
By measuring direct metrics for both approaches, we confirm that the single large kernel has lower latency and slightly higher memory usage, as shown in Figure~\ref{fig:pconv} (b). 
Furthermore, we compare the performance of the single large kernel and the stacked 3$\times$3 kernel in our small variant model.
As shown in Table ~\ref{tab:PConv}, the single large kernel has lower latency and better performance than the stacked 3$\times$3 kernel.

\begin{table}[]
\caption{Comparison of different approaches to enlarge receptive field. All metrics are measured by restoring an HD~(1280$\times$720) image using RTX4090 GPU at FP16 precision. The best results are bolded.}\label{tab:PConv}
\begin{tabular}{@{}l|cc|c@{}}
\toprule
Methods & \begin{tabular}[c]{@{}c@{}}MGO\\ (mb)\end{tabular} & \begin{tabular}[c]{@{}c@{}}Latency\\ (ms)\end{tabular} & \begin{tabular}[c]{@{}c@{}}Performance\\ (PSNR, Urban100)\end{tabular} \\ \midrule
PConv$_{3\times3}$ ($\times6$) & \textbf{226.5} & 17.2 & 32.35 \\ 
PConv$_{13\times13}$ & 228.5 & \textbf{16.5} & \textbf{32.58}  \\ \bottomrule 
\end{tabular}
\end{table}

\section{Proposed Methods}
\begin{figure*}[t!]
  \centering
  \includegraphics[width=0.8\linewidth]{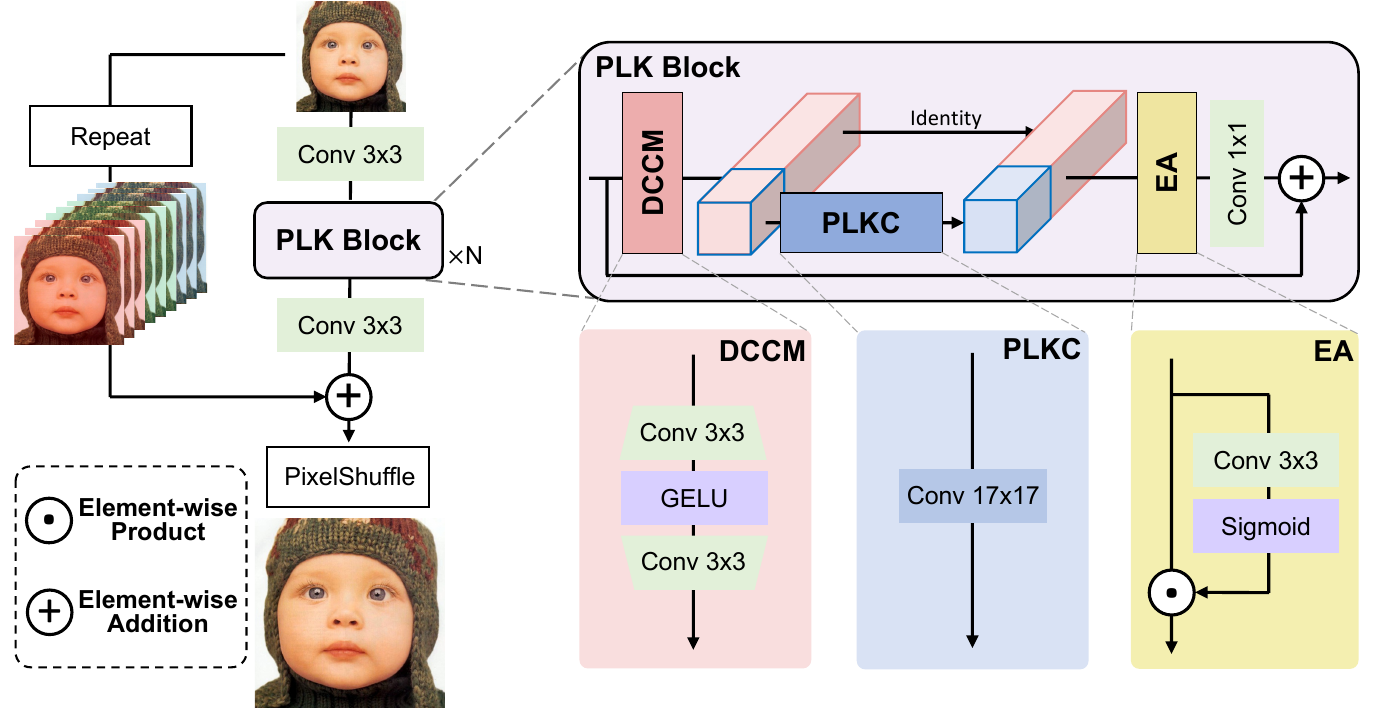}
  \caption{
    Overview of our architecture. The PLK (Partial Large Kernel) Block, the main block of PLKSR, consists of three modules: DCCM~(Double Conv Channel Mixer), PLKC~(Partial Large Kernel Conv), and EA~(Element-wise Attention). 
  }
  \label{fig:Architecture}
\end{figure*}
In this section, we demonstrate our proposed model for SR, as shown in Figure~\ref{fig:Architecture}
Initially, the low-resolution image $I^{LR}$ is processed through a 3$\times$3 convolution, followed by the N numbers of PLK Blocks sequentially, and concluded with another 3$\times$3 convolution to generate $F_{h}$.
This is formulated as Equation~\ref{eq:LearnablePath}.
\begin{equation}
\begin{gathered}
    F^{0}_{in} = \mathrm{Conv_{3\times3}}(I^{LR})\\
    F^{N}_{out} = \mathrm{PLKBlock}^{N}(~\dotsi~\mathrm{PLKBlock}^{0}(F^{0}_{in})~\dotsi~)\\
    F_{h} = \mathrm{Conv_{3\times3}}(F^{N}_{out})
    \vspace{-0.2cm}
\end{gathered}\label{eq:LearnablePath}
\end{equation}
In a parallel path, as suggested in previous work~\cite{ABPN}, the $I^{LR}$'s RGBs are repeated $\mathrm{r^{2}}$ times to produce $F_{l}$ as depicted in Equation~\ref{eq:ResidualPath}, where $r$ denotes the upscaling factor.
Following the $\mathrm{PixelShuffle}$ operation~\cite{ESPCN}, $F_{l}$ becomes the nearest interpolated image.
\begin{equation}
\begin{gathered}
    F_{l} = \mathrm{Repeat}_{r^2}(I^{LR}), \quad F_{l}\in\mathbb{R}^{3\cdot r^{2}\times h\times w}
\end{gathered}\label{eq:ResidualPath}
\end{equation}
Ultimately, the final feature map $F$ is created by adding $F_{h}$ and $F_{l}$ from each respective path, and the upscaled image $I^{SR}$ is reconstructed by applying the $\mathrm{PixelShuffle}$ operation which moves channel dimension into spatial dimension.
This is equivalent to Equation~\ref{eq:Upsample}, where $h$ and $w$ denote the feature map's height and width, respectively.
\begin{equation}
\begin{gathered}
    F = F_{h} + F_{l} \\
    I^{SR} = \mathrm{PixelShuffle}(F), \quad I^{SR}\in\mathbb{R}^{3\times h\cdot r\times w\cdot r}
\end{gathered}\label{eq:Upsample}
\end{equation}
\subsection{Partial Large Kernel Block}
The PLK Block consists of three primary modules: the Doubled Convolutional Channel Mixer~(DCCM) for local feature extraction, the Partial Large Kernel Convolution~(PLKC) to deal with long-range dependencies, and the Element-wise Attention~(EA) module for instance-dependent modulation.
The incoming feature $F_{in}$ to the PLK Block is processed sequentially through three main modules, followed by a final 1$\times$1 convolution that is subsequently added back to $F_{in}$, as illustrated in Equation~\ref{eq:PLKBlock}.
\begin{equation}
\begin{gathered}
    F_{local} = \mathrm{DCCM}(F_{in})\\
    F_{cat} = \mathrm{PLKC}(F_{local})\\
    F_{att} = \mathrm{EA}(F_{cat})\\
    F_{out} = F_{in} + \mathrm{Conv_{1\times1}}(F_{att})
\end{gathered}\label{eq:PLKBlock}
\end{equation}
\subsubsection{Double Convolutional Channel Mixer.}\label{subsubsec:ChannelMixer}
In Transformers, Feedforward Neural Networks~(FFN) are often utilized to deal with channel information. 
With $X$ and $Y$, each denoting the input and output respectively, FFN can be expressed by Equation~\ref{eq:FFN}.
The $\mathrm{MLP^{proj}}$ expands the channel's dimension($\mathbb{R}^{c\times h\times w}\rightarrow \mathbb{R}^{2\cdot c\times h\times w}$), and the $\mathrm{MLP^{agg}}$ aggregates channel information by reducing the channel to its original dimension($\mathbb{R}^{2\cdot c\times h\times w}\rightarrow \mathbb{R}^{c\times h\times w}$) where $c$ denotes feature map's channel size.
\begin{align}
\begin{split}
    Y = \mathrm{MLP^{agg}}(\mathrm{GELU}(\mathrm{MLP^{proj}}(X)))\label{eq:FFN}
\end{split}
\end{align}
Recently, several variants~\cite{SAFMN, DLGSANet, SRFormer, DAT} have been introduced, incorporating a convolution layer into FFNs to enhance high-frequency details.
Disregarding residual skips or gating mechanisms, these approaches can be fundamentally categorized into two types.
\begin{align}
    Y = \mathrm{MLP^{agg}}(\mathrm{GELU}(\mathrm{Conv^{proj}_{3\times3}}(X)))\label{eq:CCM}\\
    Y = \mathrm{Conv^{agg}_{3\times3}}(\mathrm{GELU}(\mathrm{MLP^{proj}}(X)))\label{eq:ICCM}
\end{align}
Equation~\ref{eq:CCM} corresponds to the Convolutional Channel Mixer (CCM)~\cite{SAFMN}, while Equation~\ref{eq:ICCM} introduces the concept of the Inverted Convolutional Channel Mixer~(ICCM).
Additionally, we introduce the Double Convolutional Channel Mixer (DCCM), which substitutes all MLPs with the 3$\times$3 convolutions, contrasting with existing methods that leverage the 3$\times$3 convolution only once.
\begin{align}
\begin{split}
    Y = \mathrm{Conv^{agg}_{3\times3}}(\mathrm{GELU}(\mathrm{Conv^{proj}_{3\times3}}(X))\label{eq:DCCM}
\end{split}
\end{align}
We replace all MLPs with 3x3 convolution to improve the ability to extract local features and experimentally confirm that this performs better than other channel mixing methods.
Within the PLK Block, the input $X$ corresponds to $F_{in}$, and the output $Y$ corresponds to $F_{local}$.
\subsubsection{Partial Large Kernel convolution Module.}
Upon entering the Partial Large Kernel Convolution (PLKC) module, the input feature map $F_{local}$ is divided into two feature maps based on the C channels: $F_{conv}$ for convolution and $F_{id}$ for identity. 
Subsequently, the module applies a K$\times$K large kernel convolution exclusively on $F_{conv}$, generating a large kernel-filtered feature map, denoted as $F_{global}$. 
After then, $F_{global}$ is concatenated with $F_{id}$ channel-wise, resulting in $F_{cat}$.
This is formulated as Equation~\ref{eq:PLKC}.
\begin{align}
\begin{split}
    F_{conv}, F_{id} = \mathrm{Split_{channel}}([F_{local}], C)\\
    F_{global} = \mathrm{Conv_{K\times K}}(F_{conv})\\
    F_{cat} = \mathrm{Concat_{channel}}([F_{global}, F_{id}])
\end{split}\label{eq:PLKC}
\end{align}
\begin{table*}[h]
\caption{Comparisons of the other SR methods trained on DIV2K Datasets. All metrics are measured by restoring an HD~(1280$\times$720) image using RTX4090 GPU at FP16 precision. MGO means maximum GPU memory occupancy. Performances are measured using (PSNR/SSIM) $^{\dagger}$ denotes that since their codes were unavailable, we re-implemented them. See supplement for implementation details. The best and second-best results are bolded and underlined, respectively.
}\label{tab:LightResults}
\vspace{-0.3cm}
\resizebox{\textwidth}{!}{%
\begin{tabular}{@{}l|c|cc|ccccc@{}}
\toprule
Methods & Scale & \begin{tabular}[c]{@{}c@{}}Latency\\ (ms)\end{tabular} & \begin{tabular}[c]{@{}c@{}}MGO\\ (mb)\end{tabular} & Set5 & Set14 & BSD100 & Urban100 & Manga109 \\ \midrule
EDSR\cite{EDSR} & \multirow{10}{*}{$\times$2} & 196.6 & 1488.4 & 38.11/0.9602 & 33.92/0.9195 & 32.32/0.9013 & 32.93/0.9351 & 39.10/0.9773 \\
SwinIR-light\cite{SwinIR} &  & 124.9 & 1764.3 & 38.14/0.9611 & 33.86/0.9206 & 32.31/0.9012 & 32.76/0.9340 & 39.12/0.9783 \\
ELAN-light\cite{ELAN} &  & \underline{50.2} & 687.0 & 38.17/0.9611 & 33.94/0.9207 & 32.30/0.9012 & 32.76/0.9340 & 39.11/0.9782 \\
SwinIR-NG\cite{NGSwin} &  & 143.9 & 1712.4 & 38.17/0.9612 & 33.94/0.9205 & 32.31/0.9013 & 32.78/0.9340 & 39.20/0.9781 \\
OmniSR\cite{OmniSR} &  & 76.6 & 892.2 & 38.22/\underline{0.9613} & \underline{33.98}/0.9210 & \textbf{32.36}/\textbf{0.9020} & \textbf{33.05}/\underline{0.9363} & 39.28/0.9784 \\
CRAFT\cite{CRAFT} &  & 107.2 & 632.8 & \underline{38.23}/\textbf{0.9615} & 33.92/\underline{0.9211} & \underline{32.33}/0.9016 & 32.86/0.9343 & \textbf{39.39}/\textbf{0.9786} \\
SRFormer-light\cite{SRFormer} &  & 152.7 & 1744.4 & \underline{38.23}/\underline{0.9613} & 33.94/0.9209 & \textbf{32.36}/\underline{0.9019} & 32.91/0.9353 & 39.28/\underline{0.9785} \\
DLGSANet-light\cite{DLGSANet} &  & 272.9 & \underline{562.2} & 38.20/0.9612 & 33.89/0.9203 & 32.30/0.9012 & 32.94/0.9355 & 39.29/0.9780 \\
DITN$^{\dagger}$\cite{DITN} &  & 62.8 & 679.1 & 38.17/0.9611 & 33.79/0.9199 & 32.32/0.9014 & 32.78/0.9343 & 39.21/0.9781 \\
\textbf{PLKSR~(Ours)} &  & \textbf{49.6} & \textbf{241.9} & \textbf{38.25}/\underline{0.9613} & \textbf{34.03}/\textbf{0.9214} & \textbf{32.36}/\textbf{0.9020} & \underline{32.99}/\textbf{0.9365} & \underline{39.31}/0.9781 \\ \midrule
EDSR\cite{EDSR} & \multirow{10}{*}{$\times$3} & 85.9 & 1318.9 & 34.65/0.9280 & 30.52/0.8462 & 29.25/0.8093 & 28.80/0.8653 & 34.17/0.9476 \\
SwinIR-light\cite{SwinIR} &  & 50.6 & 810.4 & 34.62/0.9289 & 30.54/0.8463 & 29.20/0.8082 & 28.66/0.8624 & 33.98/0.9478 \\
ELAN-light\cite{ELAN} &  & \underline{21.1} & 284.9 & 34.61/0.9288 & 30.55/0.8463 & 29.21/0.8081 & 28.69/0.8624 & 34.00/0.9478 \\
SwinIR-NG\cite{NGSwin} &  & 60.1 & 788.2 & 34.64/0.9293 & 30.58/\underline{0.8471} & 29.24/0.8090 & 28.75/0.8639 & \underline{34.22}/0.9488 \\
OmniSR\cite{OmniSR} &  & 27.6 & 409.3 & \underline{34.70}/0.9294 & 30.57/0.8469 & \textbf{29.28}/0.8094 & \underline{28.84}/\underline{0.8656} & \underline{34.22}/0.9487 \\
CRAFT\cite{CRAFT} &  & 43.5 & 284.7 & \textbf{34.71}/\underline{0.9295} & \textbf{30.61}/0.8469 & 29.24/0.8093 & 28.77/0.8635 & \textbf{34.29}/\textbf{0.9491} \\
SRFormer-light\cite{SRFormer} &  & 58.9 & 783.0 & 34.67/\textbf{0.9296} & 30.57/0.8469 & 29.26/\textbf{0.8099} & 28.81/0.8655 & 34.19/\underline{0.9489} \\
DLGSANet-light\cite{DLGSANet} &  & 101.8 & \underline{251.7} & \underline{34.70}/\underline{0.9295} & 30.58/0.8465 & 29.24/0.8089 & 28.83/0.8653 & 34.16/0.9483 \\
DITN$^{\dagger}$\cite{DITN} &  & 21.8 & 336.3 & 34.63/0.9290 & 30.55/0.8467 & 29.23/0.8088 & 28.68/0.8630 & 34.15/0.9482 \\
\textbf{PLKSR~(Ours)} &  & \textbf{18.5} & \textbf{131.9} & \underline{34.70}/0.9292 & \underline{30.60}/\textbf{0.8473} & \underline{29.27}/\underline{0.8096} & \textbf{28.86}/\textbf{0.8666} & 34.13/0.9485 \\ \midrule
EDSR\cite{EDSR} & \multirow{10}{*}{$\times$4} & 51.8 & 1445.5 & 32.46/0.8968 & 28.80/\underline{0.7876} & 27.71/0.7420 & 26.64/\underline{0.8033} & 31.02/0.9148 \\
SwinIR-light\cite{SwinIR} &  & 28.7 & 462.7 & 32.44/0.8976 & 28.77/0.7858 & 27.69/0.7406 & 26.47/0.7980 & 30.92/0.9151 \\
ELAN-light\cite{ELAN} &  & 18.9 & 173.5 & 32.43/0.8975 & 28.78/0.7858 & 27.69/0.7406 & 26.54/0.7982 & 30.92/0.9150 \\
SwinIR-NG\cite{NGSwin} &  & 48.5 & 450.9 & 32.44/0.8980 & 28.83/0.7870 & 27.73/0.7418 & 26.61/0.8010 & 31.09/0.9161 \\
OmniSR\cite{OmniSR} &  & 24.0 & 238.3 & 32.49/0.8988 & 28.78/0.7859 & 27.71/0.7415 & 26.64/0.8018 & 31.02/0.9151 \\
CRAFT\cite{CRAFT} &  & 43.6 & 212.3 & \underline{32.52}/0.8989 & \underline{28.85}/0.7872 & 27.72/0.7418 & 26.56/0.7995 & \textbf{31.18}/\textbf{0.9168} \\
SRFormer-light\cite{SRFormer} &  & 34.2 & 477.2 & 32.51/0.8988 & 28.82/0.7872 & \underline{27.73}/\underline{0.7422} & \underline{26.67}/0.8032 & \underline{31.17}/\underline{0.9165} \\
DLGSANet-light\cite{DLGSANet} &  & 70.3 & \underline{143.7} & \textbf{32.54}/\underline{0.8993} & 28.84/0.7871 & \underline{27.73}/0.7415 & 26.66/\underline{0.8033} & 31.13/0.9161 \\
DITN$^{\dagger}$\cite{DITN} &  & \underline{18.5} & 253.6 & \textbf{32.54}/0.8988 & \textbf{28.86}/0.7874 & 27.72/0.7420 & 26.54/0.8001 & 30.98/0.9153 \\
\textbf{PLKSR~(Ours)} &  & \textbf{10.9} & \textbf{94.4} & \textbf{32.54}/\textbf{0.8996} & 28.84/\textbf{0.7880} & \textbf{27.74}/\textbf{0.7424} & \textbf{26.69}/\textbf{0.8054} & 31.10/0.9164 \\\bottomrule
\end{tabular}
}
\vspace{-0.3cm}
\end{table*}
\subsubsection{Element-wise Attention Module.}
Element-wise Attention (EA) is an attention mechanism~\cite{PAN} that maintains the spatial and channel dimension size intact, distinguishing it from other attention mechanisms, such as spatial/channel attention, that typically involve dimensionality reduction~\cite{SENet, SCACNN, CBAM}.
We leverage this module for instance-dependent modulation similar to Transformers, as detailed in Equation~\ref{eq:PA}. 
\begin{equation}
\begin{gathered}
    F_{att} = F_{cat} \odot \mathrm{Sigmoid}(\mathrm{Conv_{3\times3}}(F_{cat}))
\end{gathered}\label{eq:PA}
\end{equation}
Empirically, we find that leveraging EA only improves performance for deeper models, so we exclude it in tiny variants.
\section{Experiments}
\subsection{Implementation Details}\label{implementation}
We implement our models using Torch 2.1.0, CUDA 12.1, and BasicSR. 
The PLKSR model consists of 28 PLK blocks and 64 channels, employing a PLKC configuration of 16 channels~(C) and a kernel size~(K) of 17. 
The PLKSR-tiny model scales down to 12 PLK blocks and 64 channels, with a PLKC of 16C and a 13K, without the PA module to leverage more blocks.

\subsection{Training Details}\label{dataset}
PLKSR is trained on DIV2K~\cite{EDSR} datasets, and two versions of PLKSR-tiny train on DIV2K and DF2K(DIV2K + Flicker2K~\cite{DF2KDataset}), respectively.
We use L1 loss to train the model using 64 patches with 96$\times$96 size, randomly flipping patches horizontally and rotating for data augmentation. 
We use the Adam~\cite{Adam} optimizer with $\beta1$ = 0.9 and $\beta2$ = 0.99 to train the model for 450k iterations. 
The initial learning rate is set as 2e-4 and subsequently halved at the [100k, 200k, 300k, 400k, 425k]-th iterations.
PLKSR$\times$3 and PLKSR$\times$4 are fine-tuned using pre-trained PLKSR$\times$2 following previous research ~\cite{EDSR, SwinIR, DITN}, while PLKSR-tiny is trained from scratch for all scales.
For fine-tuning, we use the Adam optimizer with $\beta1$ = 0.9 and $\beta2$ = 0.99 to train the model for 50k iterations. 
The initial learning rate for fine-tuning is set as 2e-4.
FP16 precision is used for all training to accelerate them.
\vspace{-0.3cm}

\begin{table*}[h]
\caption{
    Comparisons of ESR methods leveraging large kernel convolution or MHSA(-like) operation. All metrics are measured by restoring an HD~(1280$\times$720) image using RTX4090 GPU at FP16 precision. MGO means maximum GPU memory occupancy. Performances are measured using (PSNR/SSIM) $^{\dagger}$ denotes that since their codes were unavailable, we re-implemented them.  The best and second-best results are bolded and underlined, respectively.
}\label{tab:TinyResults}
\resizebox{\textwidth}{!}{%
\begin{tabular}{@{}l|c|cc|c|ccccc@{}}
\toprule
Methods & Scale & \begin{tabular}[c]{@{}c@{}}Latency\\ (ms)\end{tabular} & \begin{tabular}[c]{@{}c@{}}MGO\\ (mb)\end{tabular} & Dataset & Set5 & Set14 & BSD100 & Urban100 & Manga109 \\ \midrule
DITN-tiny$^{\dagger}$~\cite{DITN} & \multirow{6}{*}{$\times$2} & 20.2 & 569.8 & \multirow{3}{*}{DIV2K} & 38.00/0.9604 & 33.70/0.9192 & 32.16/0.8995 & 32.08/0.9282 & 38.59/0.9769 \\
DITN-real~\cite{DITN} &  & 18.8 & 543.2 & & 38.01/0.9605 & 33.65/0.9184 & 32.16/0.8995 & 31.96/0.9273 & 38.49/0.9767 \\
\textbf{PLKSR-tiny~(Ours)} &  & \underline{16.6} & \underline{228.5} & & \underline{38.11}/\underline{0.9608} & \underline{33.73}/\underline{0.9193} & \underline{32.25}/\underline{0.9008} & \underline{32.43}/\underline{0.9314} & \underline{38.84}/\underline{0.9775} \\ \cmidrule(lr){5-10}
ShuffleMixer~\cite{ShuffleMixer} &  & 28.4 & 356.5 & \multirow{3}{*}{DF2K} & 38.01/0.9606 & 33.63/0.9180 & 32.17/0.8995 & 31.89/0.9257 & 38.83/0.9774 \\
SAFMN~\cite{SAFMN} &  & \textbf{12.5} & \textbf{213.6} & & 38.00/0.9605 & 33.54/0.9177 & 32.16/0.8995 & 31.84/0.9256 & 38.71/0.9771 \\
\textbf{PLKSR-tiny~(Ours)} &  & \underline{16.6} & \underline{228.5} & & \textbf{38.14}/\textbf{0.9610} & \textbf{33.81}/\textbf{0.9199} & \textbf{32.29}/\textbf{0.9011} & \textbf{32.58}/\textbf{0.9328} & \textbf{39.18}/\textbf{0.9782} \\ \midrule
DITN-tiny$^{\dagger}$~\cite{DITN} & \multirow{6}{*}{$\times$3} & 7.4 & 284.6 & \multirow{3}{*}{DIV2K} & 34.38/0.9271 & 30.37/0.8435 & 29.10/0.8057 & 28.14/0.8529 & 33.56/0.9447 \\
DITN-real~\cite{DITN} &  & \underline{7.1} & 318.1 & & 34.33/0.9266 & 30.33/0.8424 & 29.08/0.8051 & 28.06/0.8512 & 33.44/0.9441 \\
\textbf{PLKSR-tiny~(Ours)} &  & \textbf{5.9} & \underline{103.2} & & \underline{34.50}/\underline{0.9279} & \underline{30.45}/\underline{0.8447} & \underline{29.15}/\underline{0.8070} & \underline{28.35}/\underline{0.8571} & \underline{33.71}/\underline{0.9460} \\ \cmidrule(lr){5-10}
ShuffleMixer~\cite{ShuffleMixer} &  & 10.5 & 253.2 & \multirow{3}{*}{DF2K} & 34.40/0.9272 & 30.37/0.8423 & 29.12/0.8051 & 28.08/0.8498 & 33.69/0.9448 \\
SAFMN~\cite{SAFMN} &  & \underline{7.1} & \textbf{95.73} & & 34.34/0.9267 & 30.33/0.8418 & 29.08/0.8048 & 27.95/0.8474 & 33.52/0.9437 \\
\textbf{PLKSR-tiny~(Ours)} &  & \textbf{5.9} & \underline{103.2} & & \textbf{34.54}/\textbf{0.9282} & \textbf{30.48}/\textbf{0.8455} & \textbf{29.20}/\textbf{0.8079} & \textbf{28.51}/\textbf{0.8599} & \textbf{34.05}/\textbf{0.9473} \\ \midrule
DITN-tiny$^{\dagger}$~\cite{DITN} & \multirow{6}{*}{$\times$4} & 6.5 & 248.2 & \multirow{3}{*}{DIV2K} & 32.16/0.8943 & 28.62/0.7825 & 27.59/0.7370 & 26.04/0.7853 & 30.40/0.9076 \\
DITN-real~\cite{DITN} &  & \underline{5.0} & 240.3 & & 32.10/0.8940 & 28.59/0.7813 & 27.57/0.7363 & 25.99/0.7837 & 30.29/0.9068 \\
\textbf{PLKSR-tiny~(Ours)} &  & \textbf{3.6} & \underline{65.3} & & 32.18/\underline{0.8956} & \underline{28.67}/\underline{0.7838} & \underline{27.61}/\underline{0.7380} & \underline{26.12}/\underline{0.7888} & 30.52/0.9087 \\ \cmidrule(lr){5-10}
ShuffleMixer~\cite{ShuffleMixer} & & 10.1 & 242.1 & \multirow{3}{*}{DF2K} & \underline{32.21}/0.8953 & 28.66/0.7827 & \underline{27.61}/0.7366 & 26.08/0.7835 & \underline{30.65}/\underline{0.9093} \\
SAFMN~\cite{SAFMN} &  & 7.0 & \textbf{54.7} & & 32.18/0.8948 & 28.60/0.7813 & 27.58/0.7359 & 25.97/0.7809 & 30.43/0.9063 \\
\textbf{PLKSR-tiny~(Ours)} &  & \textbf{3.6} & \underline{65.3} & & \textbf{32.33}/\textbf{0.8970} & \textbf{28.76}/\textbf{0.7857} & \textbf{27.68}/\textbf{0.7398} & \textbf{26.34}/\textbf{0.7942} & \textbf{30.83}/\textbf{0.9119} \\ \bottomrule
\end{tabular}
}
\end{table*}
\subsection{Quantitative Results}
To assess the efficiency of our SR model, we utilize latency and MGO as the main metrics.
These metrics are measured while restoring HD~(1280$\times$720) image using an RTX4090 GPU at FP16 precision.
MGO is measured by $\mathrm{\textbf{torch.cuda.max\_memory\_allocated}}$ function provided by PyTorch. 
For performance evaluation, we utilize the Peak Signal-to-Noise Ratio~(PSNR) and the Structural Similarity Index Measure~(SSIM), calculated on the Y channel in the YCbCr space after cropping the image boundary by a factor equivalent to the scaling factor. 
We assess performance across five widely used datasets: Set5~\cite{Set5}, Set14~\cite{Set14}, BSD100~\cite{BSD100}, Urban100~\cite{Urban100}, and Manga109~\cite{Manga109}.

In our comparative analysis detailed in Table~\ref{tab:LightResults}, PLKSR is evaluated with SOTA SR methods, including EDSR~\cite{EDSR}, SwinIR-light~\cite{SwinIR}, ELAN-light~\cite{ELAN}, SwinIR-NG~\cite{NGSwin}, OmniSR~\cite{OmniSR}, SRFormer-light~\cite{SRFormer}, DLGSANet-light~\cite{DLGSANet}, and DITN~\cite{DITN}. 
At a scaling factor of $\times$2, PLKSR shows similar latency ELAN-light while achieving the best PSNR on the Set5, Set14, and BSD100 datasets. 
Remarkably, against SRFormer-light on the BSD100$\times$2, PLKSR achieves the same performance with significantly reduced latency and MGO(up to 68.5\%, 86.1\% lower, respectively). 
At a scale factor of $\times$4, PLKSR outperforms ELAN-light with a 43.3\% reduction in latency, simultaneously recording the highest PSNR values on Set5, BSD100, and Urban100 datasets. 
Although some models do not use pre-training strategies, PLKSR outperforms SOTA models in both performance and efficiency, confirming PLKSR's outstanding efficiency-performance trade-off.

To highlight PLKSR's efficiency and scalability, we evaluate PLKSR-tiny with SOTA lightweight SR models employing large kernel convolution~\cite{ShuffleMixer} or MHSA(-like) mechanisms~\cite{SAFMN, DITN}.
As detailed in Table~\ref{tab:TinyResults}, PLKSR-tiny outperforms its competitors by achieving a PSNR above 39 on the Manga109 dataset at a scaling factor of $\times$2, while also showcasing the second lowest latency and MGO. 
Impressively, at a scaling factor of $\times$4, PLKSR-tiny exhibits 28\% lower latency compared to DITN-real and achieves the top performance across all evaluated datasets. 
This suggests that PLKC is scalable and is the most suitable implementation for dealing with long-range dependencies in lightweight models. 
\begin{table}[]
\caption{
    Ablation study. The KS, CM, and EA are Kernel Size, Channel Mixer, and Element-wise Attention, respectively. All metrics are measured by restoring an HD~(1280$\times$720) image using RTX4090 GPU at FP16 precision. Performances are measured using (PSNR/SSIM). The best results are bolded.
}\label{tab:ablation}
\resizebox{\columnwidth}{!}{%
\begin{tabular}{@{}cccc|cc|cc@{}}
\toprule
Blocks & KS & CM & EA & \begin{tabular}[c]{@{}c@{}}Latency\\ (ms)\end{tabular} & \begin{tabular}[c]{@{}c@{}}MGO\\ (mb)\end{tabular} & Set5 & B100 \\ \midrule
\multicolumn{1}{c|}{\multirow{5}{*}{40}} & \multicolumn{1}{c|}{5} & \multicolumn{1}{c|}{\multirow{4}{*}{CCM}} & \multirow{6}{*}{$\times$} & 36.7 & 264.5 & 38.15/0.9611 & 32.31/0.9015 \\ \cmidrule(lr){2-2}
\multicolumn{1}{c|}{} & \multicolumn{1}{c|}{9} & \multicolumn{1}{c|}{} &  & 39.0 & 267.8 & 38.18/0.9609 & 32.32/0.9017 \\ \cmidrule(lr){2-2}
\multicolumn{1}{c|}{} & \multicolumn{1}{c|}{13} & \multicolumn{1}{c|}{} &  & 43.5 & 272.9 & 38.18/0.9610 & 32.33/0.9017 \\ \cmidrule(lr){2-2}
\multicolumn{1}{c|}{} & \multicolumn{1}{c|}{\multirow{4}{*}{\textbf{17}}} & \multicolumn{1}{c|}{} &  & 48.2 & 280.0 & 38.19/0.9610 & 32.33/0.9017 \\ \cmidrule(lr){3-3}
\multicolumn{1}{c|}{} & \multicolumn{1}{c|}{} & \multicolumn{1}{c|}{ICCM} &  & 47.7 & 251.8 & 38.20/0.9612 & 32.32/0.9016 \\ \cmidrule(r){1-1} \cmidrule(lr){3-3}
\multicolumn{1}{c|}{32} & \multicolumn{1}{c|}{} & \multicolumn{1}{c|}{\multirow{2}{*}{\textbf{DCCM}}} &  & 48.1 & 241.1 & 38.21/0.9612 & 32.34/0.9018 \\ \cmidrule(r){1-1} \cmidrule(lr){4-4}
\multicolumn{1}{c|}{\textbf{28}} & \multicolumn{1}{c|}{} & \multicolumn{1}{c|}{} & \CheckmarkBold & 49.6 & 241.9 & \textbf{38.25/0.9613} & \textbf{32.36/0.9020} \\ \bottomrule
\end{tabular}
}
\end{table}

\subsection{Ablation Study}
Our ablation study evaluates the impact of kernel size, channel mixer choices, and the integration of EA.
As shown in Table~\ref{tab:ablation}, the study reveals a positive correlation between increased kernel size and improved performance, highlighting the benefits of larger convolutional kernels. 
Among various channel mixers assessed under similar latency, the DCCM emerges as the superior option, delivering the highest performance. 
Moreover, adapting the EA module achieves the most favorable balance of performance and latency, showcasing the effectiveness of our design choices in optimizing SR model efficiency. 
\begin{table}[]
\caption{
Comparison of mobile latency on ESR models leveraging large kernel convolution or MHSA(-like) operation. All latencies are measured at two input sizes using the CoreML library on iPhone~12. $^\dagger$ means that the code was not available, so we reimplemented it. $^\S$ means that the code contains an operation that cannot be converted using CoreML, so we reimplemented it. The best and second-best results are bolded and underlined, respectively.
}\label{tab:tinymobilelatency}
\resizebox{\columnwidth}{!}{
\begin{tabular}{@{}l|c|cc@{}}
\toprule
\multirow{2}{*}{Methods} & \multirow{2}{*}{Scale} & \multicolumn{2}{c}{Latency~(ms)} \\ \cmidrule(l){3-4} 
 &  & $X\in\mathbb{R}^{3\times96\times96}$ & $X\in\mathbb{R}^{3\times360\times640}$ \\ \midrule
ShuffleMixer$^{\S}$~\cite{ShuffleMixer} & \multirow{5}{*}{$\times$2} & \underline{8.6} & OOM \\
SAFMN~\cite{SAFMN} &  & 36.4 & \underline{224.3} \\
DITN-tiny$^{\dagger}$~\cite{DITN} &  & 35.2 & OOM \\ 
DITN-real~\cite{DITN} &  & 26.2 & 1460.6 \\ 
\textbf{PLKSR-tiny~(Ours)} &  & \textbf{6.0} & \textbf{181.0} \\ \bottomrule
\end{tabular}
}
\end{table}

\subsection{Comparison on Mobile device}
To validate the practicality and wide applicability of the architecture in real-world scenarios, we evaluate the latency of PLKSR-tiny on a mobile device~(iPhone 12).
We compare PLKSR-tiny with other ESR models that employ either large kernel convolution~\cite{ShuffleMixer} or MHSA(-like) mechanisms~\cite{SAFMN, DITN} on two image sizes. 
As shown in Table~\ref{tab:tinymobilelatency}, PLKSR-tiny achieves the lowest latency for both image sizes among its competitors. 
These results demonstrate that the PLKC module is the most efficient implementation for dealing with long-range dependencies on real-world edge devices.
\subsection{Large Kernel Analysis}
\begin{figure}[]
  \centering
  \includegraphics[width=\columnwidth]{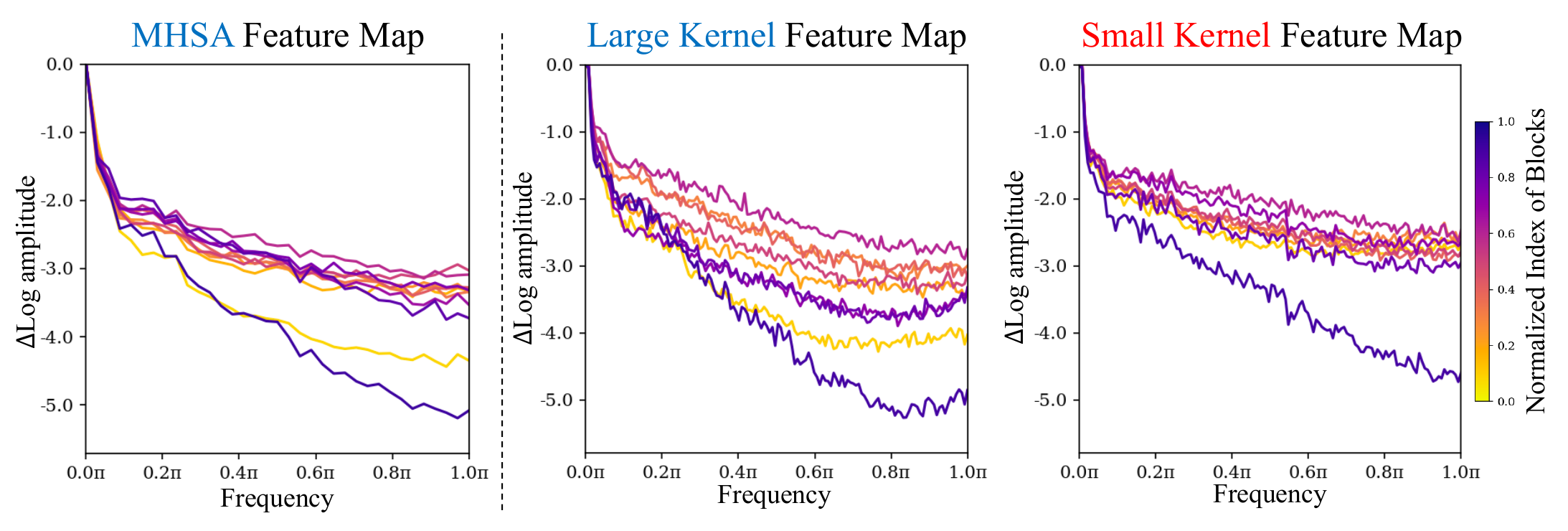}
  \caption{
    $\Delta$Log amplitude of Fourier-transformed MHSA feature maps of SRFormer-light$\times$2 and large/small kernel feature maps of PLKSR$\times$2. We visualize diagonal values after the center of each Fourier-transformed feature map following previous research~\cite{HowDoViTWorks}.
  }
  \label{fig:frequency}
\end{figure}

MHSA effectively captures low-frequency features, such as shapes and edges, in contrast to convolution, which captures high-frequency textures~\cite{HowDoViTWorks}.
To validate this, we visualize the relative log amplitude after Fourier-transforming the MHSA feature maps of SRFormer-light$\times$2 and the large/small kernel feature maps of PLKSR$\times$2.
In Figure~\ref{fig:frequency}, It can be shown that large kernels lean towards low-frequency features similar to MHSA, while small kernels capture high-frequency features in contrast to the MHSA and large kernel.

\begin{figure}[]
  \centering
  \includegraphics[width=\columnwidth]{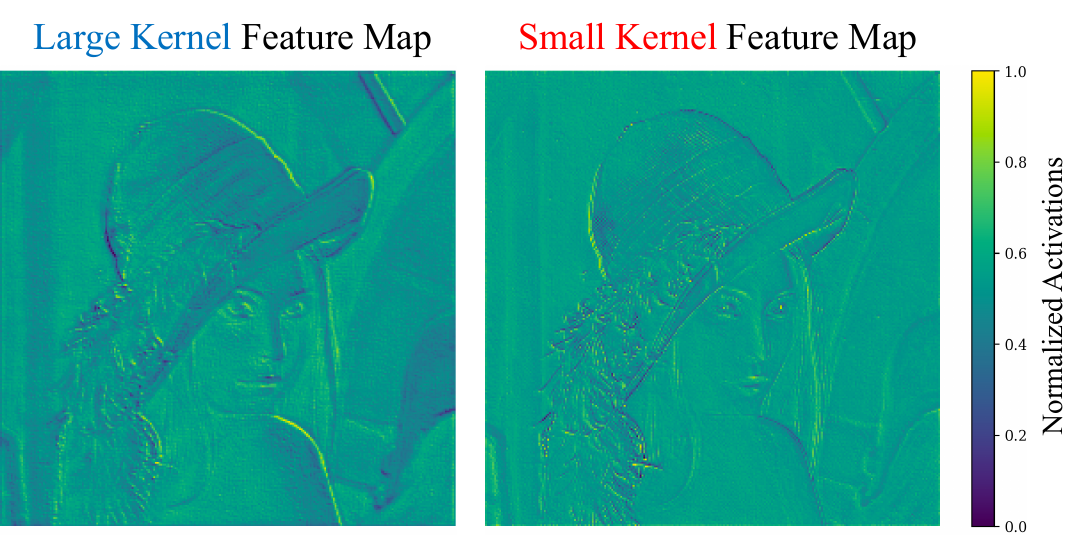}
  \caption{
    Feature map activation visualization. The large/small kernel feature maps of the last PLK block from PLKSR$\times$2 are averaged channel-wise and normalized for visualization.
  }
  \label{fig:featuremap_vis}
\end{figure}

Further visualization in Figure~\ref{fig:featuremap_vis} highlights the large kernel's ability to capture structural features, such as facial contours, and the small kernel's ability to capture textural details found in objects, such as hair or hat. 
This means that PLKC can easily capture features that are difficult to capture with the small kernel, similar to MHSA. 
This explains PLKSR's outstanding performance for the complementary use of two features.

\begin{figure}[]
  \centering
  \includegraphics[width=\columnwidth]{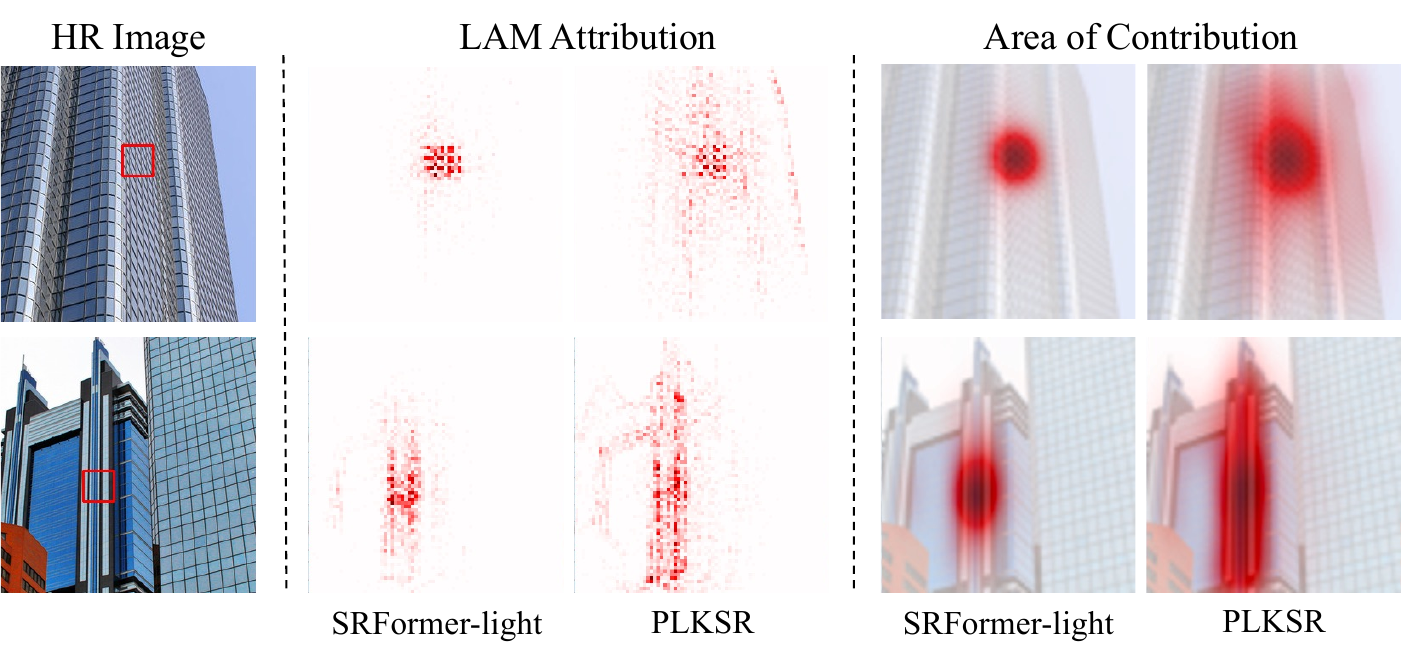}
  \vspace{-0.5cm}
  \caption{
    LAM~\cite{LAM} results of SRFormer-light~\cite{SRFormer} and PLKSR on challenging examples at Urban100$\times$4. LAM attribution represents the range of pixels used to restore the red bounding box patch, and the area of contribution represents the density of the LAM. 
  }
  \label{fig:LAM}
  \vspace{-0.5cm}
\end{figure}

\subsection{Comparison on LAM}
To demonstrate that PLKSR practically utilizes long-range dependencies captured by PLKC to reconstruct images, we introduce LAM~\cite{LAM}, a tool to represent the range of pixels that the model utilizes to reconstruct a specific region and compare it to SRFormer-light~\cite{SRFormer}.
Our model uses a 17$\times$17 large kernel, while SRFormer-light uses a 16$\times$16 window MHSA, which is a similar receptive field for comparison.
As shown in Figure~\ref{fig:LAM}, PLKSR utilizes a wider range of pixels to reconstruct the red bounding box than SRFormer-light.
Surprisingly, PLKSR utilizes almost all pixels in regions adjacent to the patch while it captures mostly edges in regions far away from the patch, which is consistent with previous experiments that demonstrated PLKC's structural bias.

\begin{figure*}[t!]
  \centering
  \includegraphics[width=\linewidth]{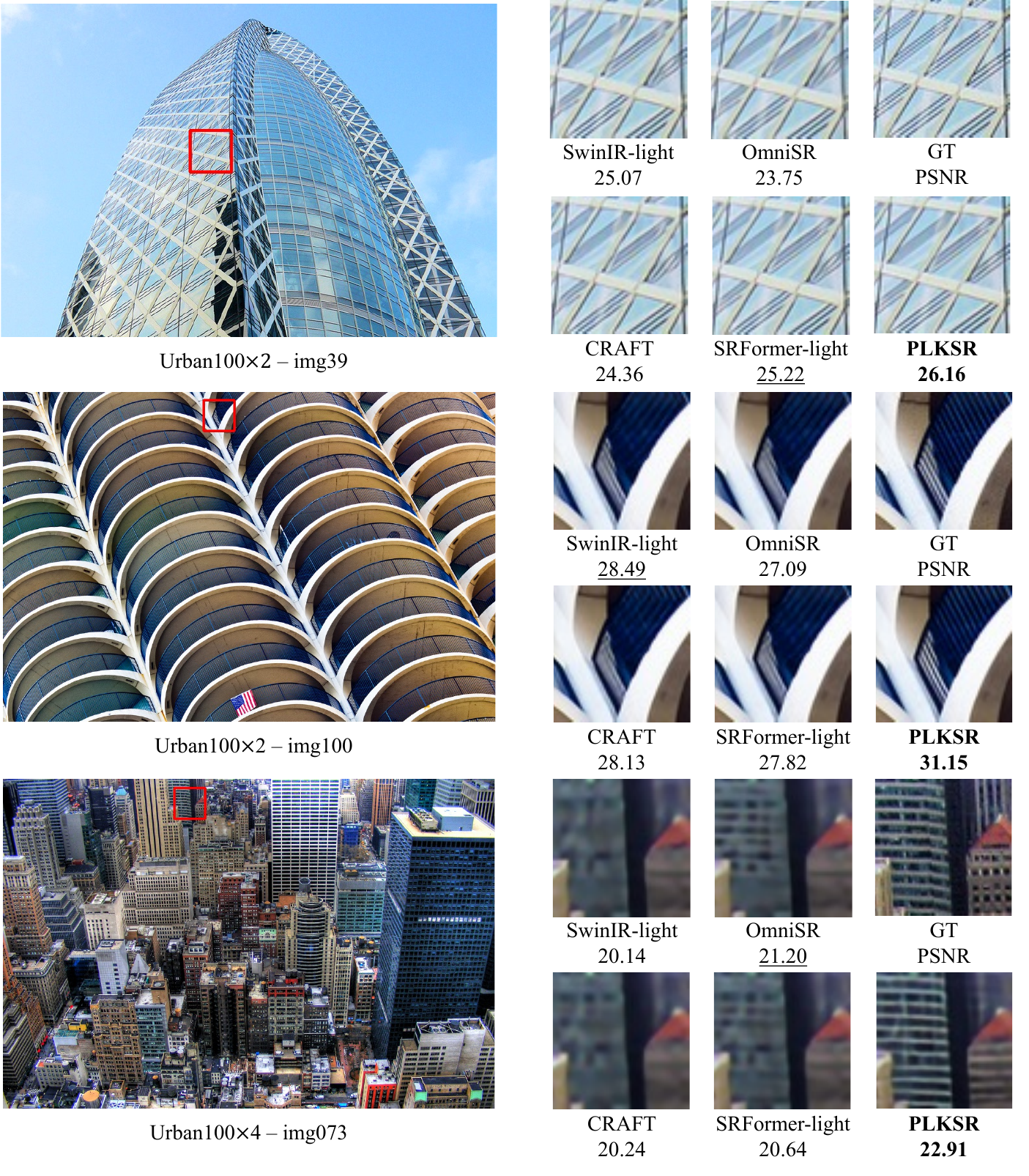}
  \vspace{-0.5cm}
  \caption{
    Visual comparisons with the other SR models on the Urban100 dataset. Visualize the result of upscaling the red bounding box and include PSNR.  The best and second-best results are bolded and underlined, respectively.
  }
  \label{fig:qualitative}
\end{figure*}
\subsection{Visual Results}
To illustrate the improved visual quality achieved by PLKSR, we visually compare the upscaled results with the other SOTA SR models, SwinIR-light~\cite{SwinIR}, OmniSR~\cite{OmniSR}, CRAFT~\cite{CRAFT}, and SRFormer-light~\cite{SRFormer} on Urban100 dataset.
In Figure~\ref{fig:qualitative}, a detailed zoom on the areas within the red bounding boxes showcases the precision with which each model performs upscaling alongside the PSNR for these reconstructions. 
Dissimilar to competitors, which often yield over-smoothed textures or fail to capture details, PLKSR successfully reconstructs the image and captures the edge.
This strongly supports the fact that PLKSR produces more visually pleasing results than other SR models.
\section{Conclusion} \label{conclusion}
This study demonstrated that contrary to the common belief that CNNs are inefficient compared to transformers, they are actually much more efficient based on direct metrics. 
We reduced the computational overhead by incorporating the advantages of transformers into CNNs, utilizing 17$\times$17 convolution for large receptive fields and element-wise attention for context-dependent weights.
PLKSR achieved state-of-the-art performance at scale $\times$4 on four datasets, outperforming ELAN-light with a 42\% reduction in latency. 
Our experiments also showed that PLKC performed similarly to MHSA on modern GPUs and edge devices, but was significantly more efficient. 
We believe PLKC will spark a comeback for CNNs in SR tasks, which currently receive less attention than transformers.

\small
\bibliographystyle{unsrt}
\bibliography{refs}
\twocolumn[
\centering
\Large
\textbf{Partial Large Kernel CNNs for Efficient Super-Resolution}

\textbf{------ Supplementary Material ------} \\
\vspace{1.0em}

\appendix

]

This supplementary material presents implementation details, results of structural re-parameterization, comparisons of GPU-optimized implementations, and additional comparisons not included in the manuscript.

\section{Implementation Details}
This section describes the implementation details.

\subsubsection{DITN}
As DITN~\cite{DITN} only released the code for DITN-real, we re-implement DITN-tiny and DITN based on the descriptions in the paper and the available code.
We re-implement DITN-tiny, replacing DITN-real's Tanh and Conv1$\times$1 with Layer Normalization according to Equation~(7) in the paper.
We also re-implement DITN by tripling the number of UPONE unit in DITN-tiny.

\subsubsection{DLGSANet}
DLGSANet's~\cite{DLGSANet} Dynamic Convolution~\cite{DWNet} is excluded from Automatic Mixed Precision due to its lack of support for FP16 precision. Consequently, we have doubled the number of CUDA threads.
As the paper reports performance using the Test-time Local Converter~(TLC)~\cite{TLC}, we also report metrics using TLC.

\subsubsection{Mobile Conversion}
Since PyTorch's \textbf{Tensor.var} function cannot be converted by CoreML, we implemente it from scratch.
Additionally, since CoreML does not convert the dynamic value assignment and \textbf{torch.repeat\_interleave} functions used by PLKSR, we re-implement these using \textbf{split}/\textbf{concatenate} and \textbf{reshape}/\textbf{permute}.

\section{Structural Re-parameterizations}
To explore performance improvements, we consider structural re-parameterization.
We compare the performances of various structural re-parameterizations~\cite{RepLKNet, SLaK, UniRepLKNet} utilized with large kernels.
However, as shown in Table~\ref{tab:RepCompare}, the optimal performance occurs without structural re-parameterization.
As these methods usally integrate with DWC in high-level vision tasks, we attribute this result to influences from different domains and large kernel approaches.

\section{GPU-optimized implementations}
In this section, we compare recently proposed GPU-optimized implementations designed to address long-range with the Partial Large Kernel Convolution~(PLKC) utilized by PLKSR.
\begin{table}[]
\caption{
    Comparisons on Structural Re-parameterization Methods. After training, kernels are merged into a single kernel, and its performance metrics~(PSNR/SSIM) are measured.
}\label{tab:RepCompare}
\vspace{-0.2cm}
\resizebox{\columnwidth}{!}{%
\begin{tabular}{@{}cc|cc@{}}
\toprule
Kernel sizes & Strides & Set5 & Urban100 \\ \midrule
{[}17$\times$17, 5$\times$5{]} & {[}1, 1{]} & 38.25/0.9613 & 32.96/0.9363 \\
{[}17$\times$5, 5$\times$17, 5$\times$5{]} & {[}1, 1, 1{]} & 38.21/0.9612 & 32.97/0.9363 \\
{[}17$\times$17, 5$\times$5, 9$\times$9, 5$\times$5, 5$\times$5{]} & {[}1, 1, 2, 3, 4{]} & 38.25/0.9614 & 32.95/0.9361 \\
\textbf{17$\times$17} & \textbf{1} & 38.25/0.9613 & 32.99/0.9365  \\ \bottomrule
\end{tabular}
\vspace{-0.2cm}
}
\end{table}

\subsection{Depth-Wise Convolution}
In recent high-level vision tasks, several studies have employed large kernels based on inverse implicit Generalized Matrix Multiplication~(iGEMM) for Depth-Wise Convolution~(DWC)~\cite{RepLKNet, SLaK, UniRepLKNet}, as also noted in a previous Super-Resolution~(SR) study~\cite{ShuffleMixer}.
While exploring various implementations for large kernels, we discover that although they prove efficient for general high-level vision tasks, they do not yield the same efficiency for SR tasks.
Table~\ref{tab:DWCCompare} shows that iGEMM DWC is significantly slower than Pytorch’s standard DWC.
This discrepancy is due to iGEMM DWC being optimized for high throughput at large batch sizes, whereas Pytorch's DWC is better suited for SR tasks processing feature maps with small batch sizes and large spatial sizes.
Consistently, the iGEMM DWC shows a minimal increase in latency as the batch size increases, while the Pytorch DWC shows a significant increase in latency.
Notably, PLKC is the most efficient, offering the lowest latency and Maximum GPU-memory Occupancy(MGO) across all batch sizes.

\begin{table}[]
\caption{
    Comparisons on Depth-wise Convolutions and PConv. 
    Metrics are calculated using a 17$\times$17 kernel on a feature map $F\in\mathbb{R}^{64\times640\times360}$ via RTX4090 GPU at FP16 precision. PLKC only processes the first 16 channels of the feature map.
}\label{tab:DWCCompare}
\vspace{-0.3cm}
\begin{tabular}{@{}c|l|cc@{}}
\toprule
Batch Size & Methods & Latency~(ms) & MGO~(mb) \\ \midrule
\multirow{3}{*}{1} & Pytorch DWC & 0.6 & 35.3 \\
 & iGEMM DWC & 318.4 & 35.3 \\ 
 & \textbf{PLKC~(16C)} & 0.2 & 44.5 \\ \cmidrule(lr){1-4}
\multirow{3}{*}{64} & Pytorch DWC & 37.5 & 2250.1 \\
 & iGEMM DWC & 324.7 & 2250.1 \\ 
 & \textbf{PLKC~(16C)} & 7.4 & 2812.9 \\\bottomrule
\end{tabular}
\vspace{-0.3cm}
\end{table}

\begin{table*}[h]
\caption{
    Comparisons on Various Large Kernel Approaches. All metrics are measured by restoring an HD~(1280$\times$720) image using RTX4090 GPU at FP16 precision. MGO means maximum GPU memory occupancy. The best results are bolded.
}\label{tab:LKCompare}
\vspace{-0.2cm}
\resizebox{\textwidth}{!}{%
\begin{tabular}{@{}l|c|cc|ccccc@{}}
\toprule
Methods & Scale & \begin{tabular}[c]{@{}c@{}}Latency\\ (ms)\end{tabular} & \begin{tabular}[c]{@{}c@{}}MGO\\ (mb)\end{tabular} & Set5 & Set14 & BSD100 & Urban100 & Manga109 \\ \midrule
Dynamic Conv~\cite{DLGSANet} & \multirow{3}{*}{$\times$2} & 55.4 & 2216.2 & 37.67/0.9592 & 33.20/0.9145 & 31.94/0.8968 & 31.10/0.9177 & 37.61/0.9748 \\
Depth-wise Conv &  & 51.1 & 216.4 & 38.18/0.9611 & 33.81/0.9198 & 32.31/0.9014 & 32.61/0.9335 & 39.00/0.9777 \\
\textbf{PLKC} &  & 49.6 & 241.9 & \textbf{38.25}/\textbf{0.9613} & \textbf{34.03}/\textbf{0.9214} & \textbf{32.36}/\textbf{0.9020} & \textbf{32.99}/\textbf{0.9365} & \textbf{39.31}/\textbf{0.9781} \\ \bottomrule
\end{tabular}
}
\end{table*}
\begin{table*}[h]
\caption{
    Comparisons of Partial Channel Designs. All metrics are measured by restoring an HD~(1280$\times$720) image using RTX4090 GPU at FP16 precision. MGO means maximum GPU memory occupancy. The best results are bolded.
}\label{tab:PSASR}
\vspace{-0.2cm}
\resizebox{\textwidth}{!}{%
\begin{tabular}{@{}l|c|cc|ccccc@{}}
\toprule
Methods & Scale & \begin{tabular}[c]{@{}c@{}}Latency\\ (ms)\end{tabular} & \begin{tabular}[c]{@{}c@{}}MGO\\ (mb)\end{tabular} & Set5 & Set14 & BSD100 & Urban100 & Manga109 \\ \midrule
Partial ASA~\cite{ELAN} & \multirow{2}{*}{$\times$2} & 50.7 & 568.5 & 38.23/\textbf{0.9613} & 33.98/0.9211 & 32.34/0.9018 & 32.87/0.9355 & 39.16/0.9780 \\
\textbf{PLKC} &  & 49.6 & 241.9 & \textbf{38.25}/\textbf{0.9613} & \textbf{34.03}/\textbf{0.9214} & \textbf{32.36}/\textbf{0.9020} & \textbf{32.99}/\textbf{0.9365} & \textbf{39.31}/\textbf{0.9781} \\ \bottomrule
\end{tabular}
}
\end{table*}

\subsection{Flash Attention}
Flash Attention has recently been introduced to enhance the memory access pattern of Multi-Head Self-Attention (MHSA), thus accelerating inference speed and reducing MGO on GPU devices.
To demonstrate the superiority of PLKC over the most optimized MHSA implementations, we assess the latency and MGO of various SR Transformers~\cite{SwinIR, SRFormer} accelerated by Flash Attention-2~\cite{flashattn}.
We exclude attention masks and relative positional bias for shifted windows due to their unavailability.
Table~\ref{tab:flashattn} illustrates that, at a scaling factor of $\times$4, SwinIR${_\mathrm{F}}$-light, when accelerated by Flash Attention-2, shows a 49\% reduction in latency compared to the original SwinIR-light.
Notably, this result underscores the exceptional efficiency of PLKC since PLKSR remains 25\% faster than SwinIR${_\mathrm{F}}$-light, despite this comparison being the highly favorable setting for Transformers.

\begin{table}[]
\caption{
    Comparison of Transformers accelerated by Flash Attention. $_\mathrm{F}$ denotes each model is converted to use Flash Attention-2~\cite{flashattn}. All metrics are measured by restoring an HD~(1280$\times$720) image using RTX4090 GPU at FP16 precision. The best results are bolded.
}
\label{tab:flashattn}
\vspace{-0.2cm}
\resizebox{\columnwidth}{!}{%
\begin{tabular}{@{}l|ccc@{}}
\toprule
\multirow{2}{*}{Methods} & \multicolumn{3}{c}{Latency~(ms) / MGO~(mb)} \\ \cmidrule(l){2-4} 
                         & $\times$2        & $\times$3       & $\times$4       \\ \midrule
SwinIR$_{\mathrm{F}}$-light & 68.0 / 1235.9 & 24.9 / 573.4 & 14.6 / 328.6 \\
SRFormer$_{\mathrm{F}}$-light& 89.7 / 1281.7 & 31.1 / 581.1 & 19.6 / 356.6 \\
\textbf{PLKSR~(Ours)} & \textbf{49.6} / \textbf{241.9} & \textbf{18.5} / \textbf{131.9} & \textbf{10.9} / \textbf{94.4} \\ \bottomrule
\end{tabular}
\vspace{-0.2cm}
}
\end{table}

\section{Comparison on other LK approaches}
Recent studies have explored using large kernels in SR tasks, leveraging approaches like Depth-wise Convolution~\cite{ShuffleMixer} and Dynamic Convolution~\cite{DLGSANet}.
To demonstrate PLKC's superiority, we substitute PLKC with other large kernel approaches and compare their performance.
We adjust the number of main blocks in each model to equalize latency, and we omit Element-wise Attention~(EA) since Dynamic Convolution has instance-dependent weights.
As indicated in Table~\ref{tab:LKCompare}, PLKC not only achieves the lowest latency but also delivers the best results, thereby confirming its remarkable performance.

\section{Comparison on Self-Attention}
SHViT~\cite{SHViT} has achieved state-of-the-art performance and latency by computing single-head self-attention on a subset of channels.
Given the similarity of this approach to ours in dealing with long-range dependencies on partial channels, we substitute PLKC with ELAN's Accelerated Self-Attention (ASA)~\cite{ELAN} with partial channel design for performance comparison.
We adjust the number of main blocks to equalize latency and omit Element-wise Attention~(EA) since ASA has instance-dependent weights.
As demonstrated in Table~\ref{tab:PSASR}, PLKSR outperforms ASA-variants across all datasets, thereby underscoring PLKC's outstanding efficiency and performance.
\begin{figure}[]
  \centering
  \includegraphics[width=\columnwidth]{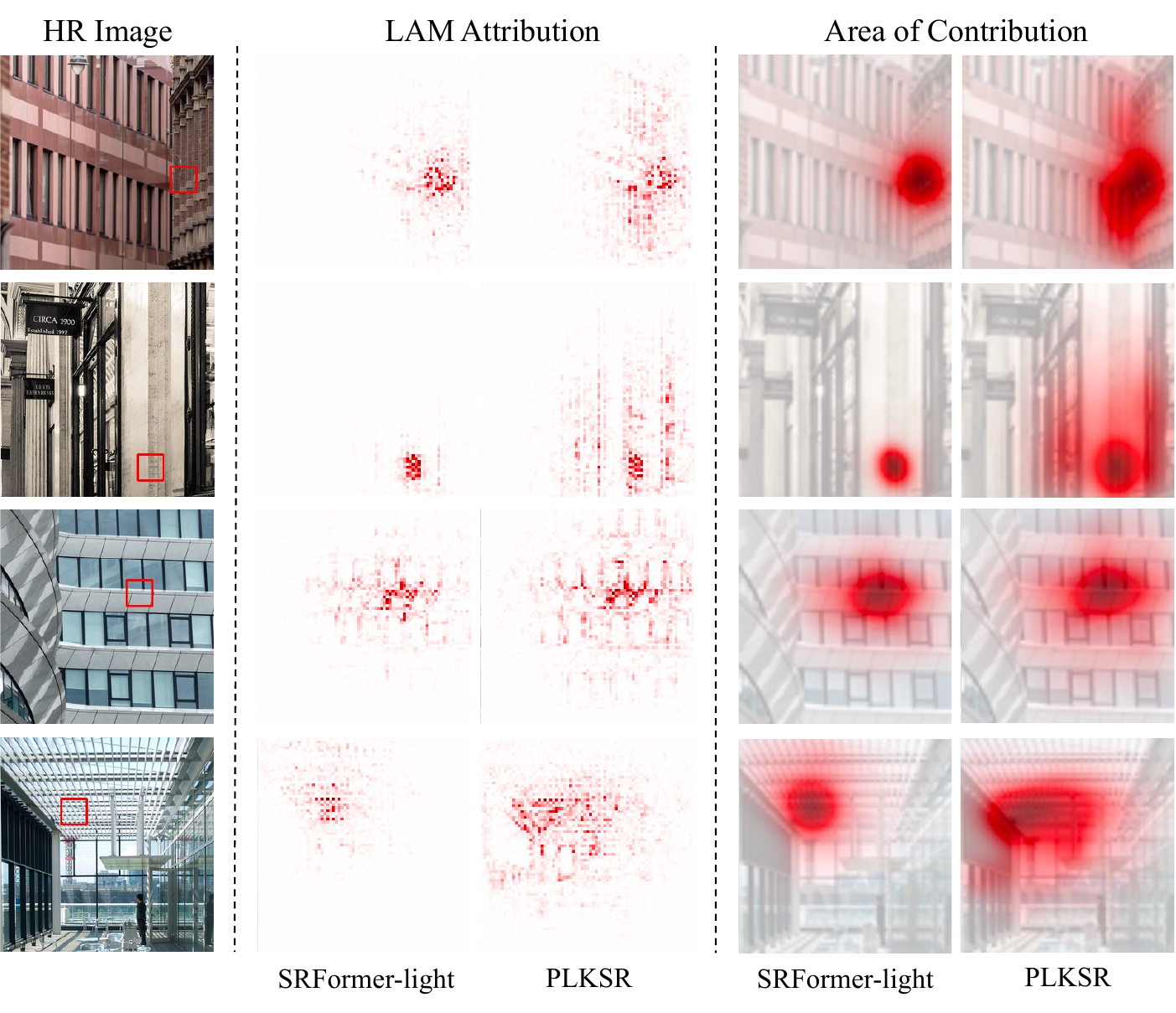}
  \vspace{-0.7cm}
  \caption{
  More LAM results of SRFormer-light~\cite{SRFormer} and PLKSR on challenging examples at Urban100×4. LAM attribution represents the range of pixels used to restore the red bounding box patch, and the area of contribution represents the density of the LAM.
  }
  \label{fig:MoreLAM}
  \vspace{-0.3cm}
\end{figure}

\section{More Comparison on LAM}
In this section, we present additional Local Attention Map~(LAM)~\cite{LAM} results with our PLKSR and SRFormer-light~\cite{SRFormer}.
The figure~\ref{fig:MoreLAM} shows that PLKSR utilizes a wider range of pixels compared to SRFormer-light in various examples, demonstrating that PLKSR effectively utilizes long-range dependent features captured by PLKC.

\section{More Visual Results}
\begin{figure*}[t!]
  \centering
  \includegraphics[width=\linewidth]{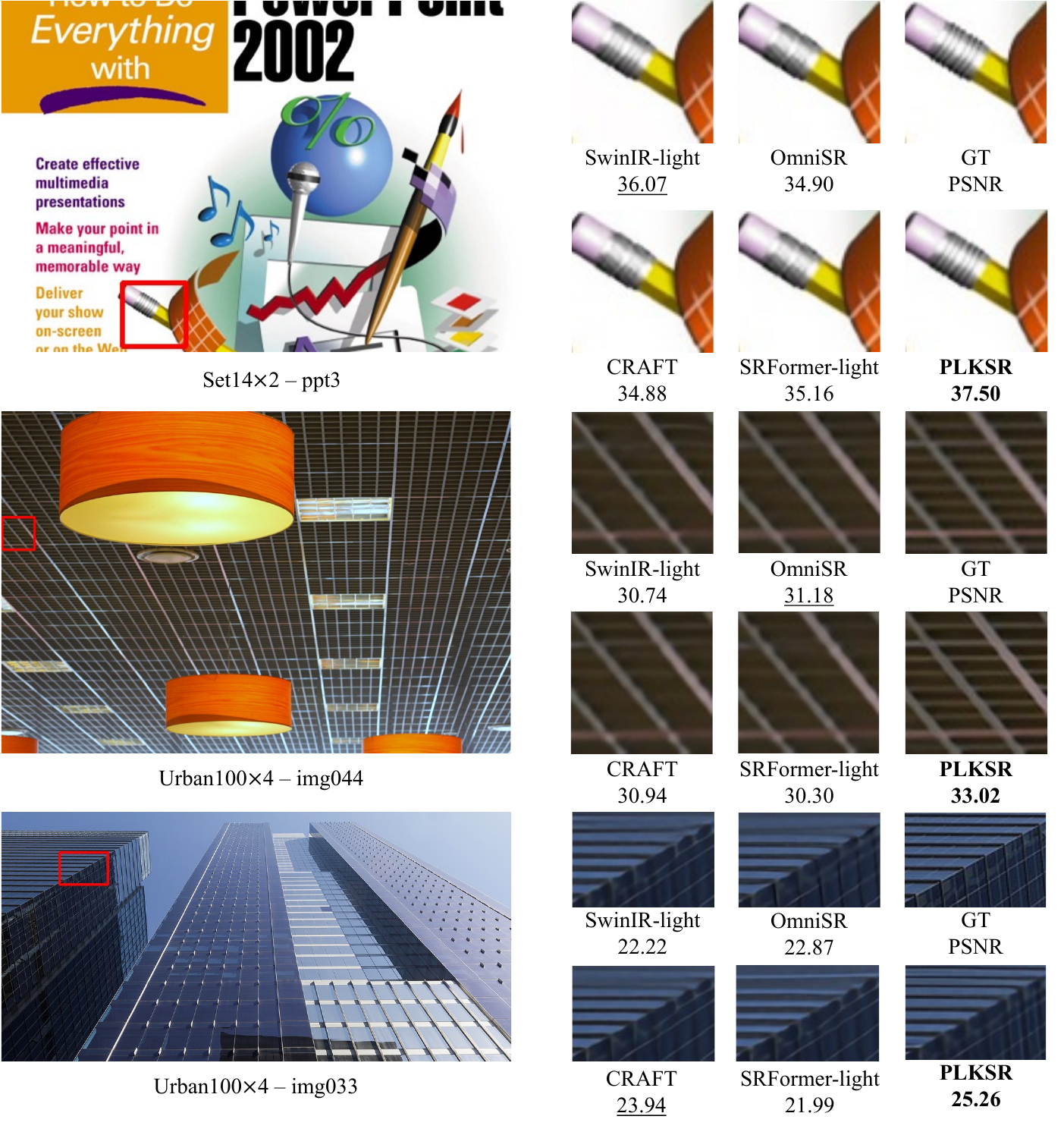}
  \caption{
    More visual results on Set14 and Urban100 datasets. The best and second-best results are bolded and underlined, respectively.
  }
  \label{fig:QulitativeSupp}  
\end{figure*}

This section provides additional visual evaluations of our PLKSR alongside other state-of-the-art methods.
As demonstrated in Figure~\ref{fig:QulitativeSupp}, PLKSR not only achieves a higher PSNR but also delivers superior visual results, surpassing other models.

\end{document}